\documentclass[journal]{IEEEtran}

\usepackage[utf8]{inputenc}
\usepackage[T1]{fontenc}
\usepackage{microtype} 
\usepackage{cite}
\usepackage{multirow}
\usepackage{graphicx}
\usepackage[caption=false]{subfig}
\usepackage{amsfonts,amsmath,amsthm,amssymb}
\usepackage{hyperref}
 \usepackage{cite,graphicx,lipsum}
 \usepackage{algorithm} 
\usepackage{algpseudocode}
 \usepackage{cite}

\usepackage{tikz}
\usetikzlibrary{automata,positioning}
\usetikzlibrary{decorations.pathreplacing,decorations.markings,shapes.geometric}
\usetikzlibrary{shapes,arrows}
\usetikzlibrary{backgrounds,calc,positioning}

\usepackage{tabularx}       
\usepackage{float}          
\usepackage{booktabs}       

\usepackage{enumerate}

\def\be{ \begin{equation} }
\def\ee{ \end{equation} }
\def\bea{ \begin{eqnarray} }
\def\eea{ \end{eqnarray} }

\def\b0{{\bf 0}}

\catcode`,\active

\catcode`\,12

\theoremstyle{remark}

\usepackage{listings}

\usepackage{xcolor}

\definecolor{codegreen}{rgb}{0,0.6,0}
\definecolor{codegray}{rgb}{0.5,0.5,0.5}
\definecolor{codepurple}{rgb}{0.58,0,0.82}
\definecolor{backcolour}{rgb}{0.95,0.95,0.92}
\lstdefinestyle{mystyle}{
  backgroundcolor=\color{backcolour}, commentstyle=\color{codegreen},
  keywordstyle=\color{magenta},
  numberstyle=\tiny\color{codegray},
  stringstyle=\color{codepurple},
  basicstyle=\ttfamily\footnotesize,
  breakatwhitespace=false,         
  breaklines=true,                 
  captionpos=b,                    
  keepspaces=true,                 
  numbers=left,                    
  numbersep=5pt,                  
  showspaces=false,                
  showstringspaces=false,
  showtabs=false,                  
  tabsize=2
}

\usepackage{enumerate}
 \usepackage[switch,pagewise]{lineno}
 \usepackage{hyperref}
 \hypersetup{
       colorlinks = true,
        citecolor=red,
 }

\def\be{ \begin{equation} }
\def\ee{ \end{equation} }
\def\bea{ \begin{eqnarray} }
\def\eea{ \end{eqnarray} }

\def\b0{{\bf 0}}

\begin{document}
\title{Modeling Quantum Machine Learning for Genomic Data Analysis}



\author{Navneet Singh 
        and
        Shiva Raj Pokhrel, \textit{IEEE, Senior Member}
        \thanks{N.~Singh and S.~R.~Pokhrel are with Deakin University, Geelong, VIC, Australia. Email: n.navneetsingh@deakin.edu.au; shiva.pokhrel@deakin.edu.au.}}%

\maketitle
\begin{abstract}

Quantum Machine Learning (QML) continues to evolve, unlocking new opportunities for diverse applications. In this study, we investigate and evaluate the applicability of QML models for binary classification of genome sequence data by employing various feature mapping techniques. We present an open-source, independent Qiskit-based implementation to conduct experiments on a benchmark genomic dataset. Our simulations reveal that the interplay between feature mapping techniques and QML algorithms significantly influences performance. Notably, the Pegasos Quantum Support Vector Classifier (Pegasos-QSVC) exhibits high sensitivity, particularly excelling in recall metrics, while Quantum Neural Networks (QNN) achieve the highest training accuracy across all feature maps. However, the pronounced variability in classifier performance, dependent on feature mapping, highlights the risk of overfitting to localized output distributions in certain scenarios. This work underscores the transformative potential of QML for genomic data classification while emphasizing the need for continued advancements to enhance the robustness and accuracy of these methodologies.


\end{abstract}

\section{Introduction}

The development of Noisy Intermediate-Scale Quantum (NISQ) devices, harnessing quantum information and physics, marks an epoch in computing \cite{preskill2018quantum}. These quantum devices promise applications beyond classical computers, positioning quantum machine learning (QML) as a key research area \cite{gil2024understanding}. This drive, fueled by the exponential power of quantum machines and advances in QML, radically alters large scale data processing. In recent years, QML have shown multiple use-cases in different domains such finance, astronomy, automobile and healthcare. However, finding practically important problems for quantum hardware and algorithms to work remains challenging \cite{gentinetta2024complexity}. As noted in \cite{flother2023state}, integrating QML into healthcare can revolutionize genomic analysis by leveraging the computational power of quantum algorithms. 
\begin{figure*}[]
\begin{center}
\includegraphics[scale=0.999]{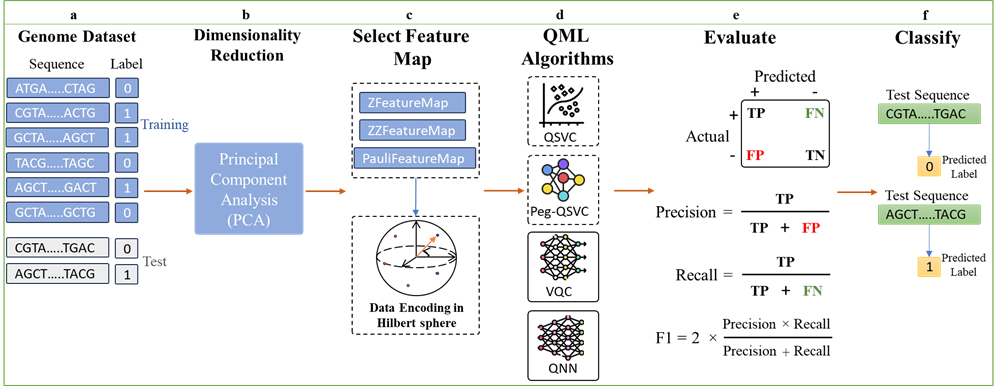}
\end{center}
\caption{\rm Illustration of the proposed workflow in this paper. We outline a method for applying QML techniques to classical Genomic datasets using NISQ devices: \textbf{a)} Dataset Split: Divide the classical dataset into training and test sets.
\textbf{b)} Dimensionality Reduction: Reduce the dataset to four dimensions using Principal Component Analysis (PCA) due to NISQ device limitations.
\textbf{c)} Quantum Encoding: Encode the dataset into quantum data using ZFeatureMap, ZZFeatureMap, and PauliFeatureMap for Hilbert space representation.
\textbf{d} QML Training: Train various QML algorithms on the quantum data.
\textbf{e)} Performance Metrics: Evaluate true positives (TP), false positives (FP), true negatives (TN), and false negatives (FN) to calculate accuracy, precision, recall, and F1 score. 
\textbf{f)} Classification: Use the trained model to classify test sequences.
}
        \label{Overall_flow}
\end{figure*}
In genomics, classical ML techniques have led to significant advancements in understanding genetic disorders, evolutionary biology, and personalized medicine by handling large, complex datasets\cite{ravi2016deep}. However,  ML methods struggle to handle the high-dimensional, complex nature of genomic data, typically necessitating sophisticated models that are computationally expensive and lack scalability \cite{ching2018opportunities,10415102,zou2019primer}. 

In this paper, we advocate that QML approaches enable the utilization of quantum computing power to analyze complex genomic data at a scale more efficiently than that of ML~\cite{pokhrel2024quantum, 10.1145/3663408.3665806}. We seek to develop a framework to serve genome classification well, where genomic sequences are analyzed to identify genetic variation (function), evolutionary relationships, and disease associations. Such a genome classification is very important for the diagnosis and treatment of genetic disorders, evolutionary biology studies, and personalized medicine. We explain and demonstrate how QML is poised to help us overcome the challenges of increasingly available genomic data based on a complex structure on a large scale while maintaining a fast computational classification algorithm \cite{ullah2024quantum, 10.1145/3663408.3665808}.

Despite significant advances in QML for medical imaging analysis, genomic sequence classification remains an underexplored frontier. Our research addresses this gap by critically assessing, expanding, and reevaluating baseline performance on genomic data \cite{pokhrel2024quantum} using prominent QML models, including the Quantum Support Vector Classifier (QSVC), Pegasos-QSVC \cite{gentinetta2024complexity}, Variational Quantum Classifier (VQC), and Quantum Neural Network (QNN) \cite{abbas2021power}. Another focus point of this work is to investigate the impact of various data encoding/feature mapping techniques, which transform genomic sequences into quantum states, on the performance of these QML models in classification tasks. By advancing QML’s application to genomics, this study seeks to bridge the gap between theoretical advancements and practical implementations, driving progress in both scientific understanding and medical innovation. 

\subsection{Motivating Research Questions}
How can QML algorithms be scaled and optimized to handle the complexity and volume of genomic data effectively? Furthermore, what strategies can mitigate the constraints of current NISQ devices, such as limited coherence times and gate fidelities? Addressing these challenges is essential for establishing a robust framework to benchmark QML algorithms against classical methods in genomic data analysis. In addition, how can quantum algorithms seamlessly integrate with classical data preprocessing techniques for efficient genomic data analysis workflows? Finally, what are the practical benefits of QML for genomic sequence classification and how can they be effectively demonstrated?

\subsection{Our Proposed Solution Approach and Contribution}
In this paper, we address the questions as follows.

\noindent \textit{Customized QML Algorithms:} We refine and enhance QML algorithms such as QSVM, Pegasos-QSVM, VQC, and QNN to improve scalability, efficiency, and accuracy for genomic data. This includes addressing the limitations of NISQ devices by implementing dinmentionality reduction, feature mapping techniques and optimizing quantum circuits to handle large and complex datasets effectively.

\noindent \textit{Standardized Benchmarking Framework:} We establish a simple proof of concept framework to systematically compare different QML algorithms as shown in Fig. ~\ref{Overall_flow}. This framework identifies the areas where quantum methods excel, providing a transparent and replicable process to evaluate the performance of QML algorithms in genomic data analysis.

\noindent \textit{Integration of Quantum and Classical Techniques:} We design preliminary methods to integrate quantum algorithms with classical preprocessing workflows seamlessly. By leveraging the strengths of feature mapping into quantum, we maximize the efficiency and effectiveness of genomic data analysis, ensuring a smooth and robust workflow.

\noindent \textit{Theoretical Evaluations and Demonstrations:} We extend QML techniques, develop their convergence analyses and demonstrate to real-world genomic classification challenges, with their potential to revolutionize genetic disorder diagnosis, evolutionary biology research, and personalized medicine. Through rigorous empirical studies and Qiskit implementations, we showcase the practical benefits of QML in genomics and how the use of different feature mapping techniques with QML model impact the  performance matrices.

\section{Background and Implementation}

\subsection{Feature Mapping} Feature maps are essential in QML models as they encode classical data $(\vec{x}_j,y_j)$ into quantum states $\phi$ for efficient processing and analysis. This section discusses three types of feature maps: ZFeatureMap, ZZFeatureMap, and PauliFeatureMap, each employing different methods to transform and entangle qubits based on classical data \cite{havlivcek2019supervised}.

\begin{figure}[t]
\begin{center}
\includegraphics[scale=0.497]{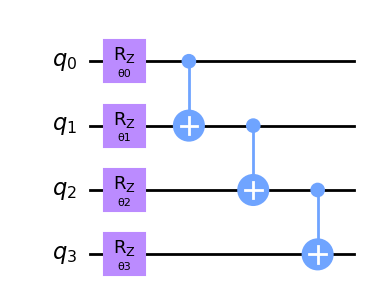}
\end{center}
\vspace{-6 mm}
\caption{\rm Quantum circuit of ZFeature Map.}
        \label{Zfmap}
        \vspace{-3 mm}
\end{figure}

\subsubsection{\textbf{ZFeatureMap}}
The ZFeatureMap \cite{schuld2020circuit} encodes classical data into quantum states by applying rotations around the Z-axis for each qubit. Given a classical data vector \(\vec{x} = (x_1, x_2, \ldots, x_n)\), each element \(x_j\) is used to rotate the \(j\)-th qubit. For a single qubit, the rotation is represented by the unitary operator \(R_z(x_j) = e^{-i x_j Z}\), where the Pauli-Z matrix \(Z\) is \(\begin{pmatrix} 1 & 0 \\ 0 & -1 \end{pmatrix}\). The matrix exponential simplifies to \(e^{-i x_j Z} = \cos(x_j) I - i \sin(x_j) Z\), resulting in \(\begin{pmatrix} e^{-i x_j} & 0 \\ 0 & e^{i x_j} \end{pmatrix}\). For \(n\) qubits, the ZFeatureMap applies this rotation to each qubit independently: \(U_Z(\vec{x}) = \bigotimes_{j=1}^n R_z(x_j)\). 
Fig.~\ref{Zfmap} illustrates the quantum circuit of ZFeatureMap, where each qubit \( q_j \) undergoes a rotation around the Z-axis by an angle \( \theta_j \), represented by the \( R_z(\theta_j) \) gate. These rotations encode the classical data vector \( \vec{x} \) into quantum states, followed by a series of CNOT gates to entangle the qubits, enhancing the expressiveness of the feature map.

\subsubsection{\textbf{ZZFeatureMap}}

\begin{figure}[h]
\begin{center}
\includegraphics[scale=0.333]{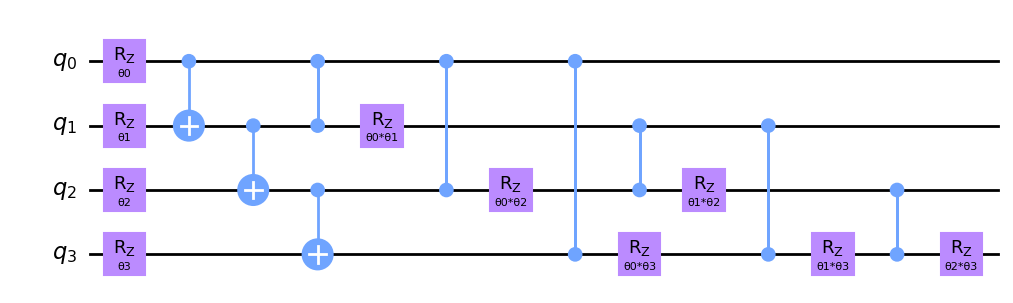}
\end{center}
\vspace{-5 mm}
\caption{\rm Quantum circuit of ZZFeature Map.}
        \label{zzfmap}
\end{figure}

The ZZFeatureMap \cite{havlivcek2019supervised} as shown in Fig.~\ref{zzfmap} extends the ZFeatureMap by incorporating interactions between qubits using controlled-Z (CZ) gates to encode pairwise feature products. The single-qubit rotation remains the same as in the ZFeatureMap. The interaction term between qubits \(j\) and \(k\) is represented by \(e^{-i x_j x_k Z_j Z_k}\), where \[Z_j \otimes Z_k = \begin{pmatrix} 1 & 0 \\ 0 & -1 \end{pmatrix} \otimes \begin{pmatrix} 1 & 0 \\ 0 & -1 \end{pmatrix}.\] The matrix exponential is \[e^{-i x_j x_k Z_j Z_k} = \begin{pmatrix} e^{-i x_j x_k} & 0 & 0 & 0 \\ 0 & e^{i x_j x_k} & 0 & 0 \\ 0 & 0 & e^{i x_j x_k} & 0 \\ 0 & 0 & 0 & e^{-i x_j x_k} \end{pmatrix}.\] The full map combines single-qubit rotations with pairwise interactions: \(U_{ZZ}(\vec{x}) = \left( \bigotimes_{j=1}^n e^{-i x_j Z_j} \right) \prod_{j < k} e^{-i x_j x_k Z_j Z_k}\) \cite{havlivcek2019supervised}. 
Fig.~\ref{zzfmap} illustrates the quantum circuit of the ZZFeatureMap, where each qubit \( q_j \) undergoes a rotation around the Z-axis by an angle \( \theta_j \), represented by the \( R_z(\theta_j) \) gate. The circuit also includes controlled-Z (CZ) gates to introduce interactions between pairs of qubits, represented by the terms \(e^{-i x_j x_k Z_j Z_k}\), which encode pairwise feature products \(\theta_i \theta_j\) into the quantum states.

\begin{figure}[t]
\begin{center}
\includegraphics[scale=0.417]{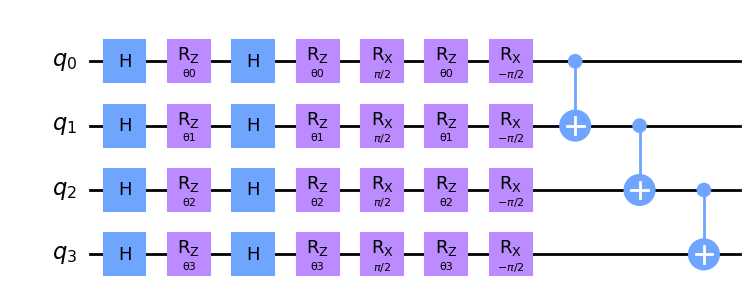}
\end{center}\vspace{-5 mm}
\caption{\rm Quantum circuit of PauliFeature Map.}
        \label{pfmap}
\end{figure}
\subsubsection{\textbf{PualiFeatureMap}}
The PauliFeatureMap \cite{havlivcek2019supervised,benedetti2019generative} 
generalizes the feature mapping by including rotations around all three axes (X, Y, Z) of the Bloch sphere and introducing more complex entanglements between qubits. 
The Pauli matrices are \(X = \begin{pmatrix} 0 & 1 \\ 1 & 0 \end{pmatrix}\), \(Y = \begin{pmatrix} 0 & -i \\ i & 0 \end{pmatrix}\), and \(Z = \begin{pmatrix} 1 & 0 \\ 0 & -1 \end{pmatrix}\). The rotations around the X, Y, and Z axes for qubit \(j\) are \(R_x(x_j) = e^{-i x_j X}\), \(R_y(x_j) = e^{-i x_j Y}\), and \(R_z(x_j) = e^{-i x_j Z}\). 
The matrix exponentials are \(R_x(x_j) = \cos(x_j)I - i \sin(x_j)X\), \(R_y(x_j) = \cos(x_j)I - i \sin(x_j)Y\), and \(R_z(x_j) = \begin{pmatrix} e^{-i x_j} & 0 \\ 0 & e^{i x_j} \end{pmatrix}\). For \(n\) qubits, the combined rotations are \[U_{Pauli}(\vec{x}) = \prod_{j=1}^n \left( R_x(x_j) R_y(x_j) R_z(x_j) \right).\] 
Entangling terms such as \(e^{-i x_j x_k Z_j X_k}\) are added to introduce entanglement, where \[Z_j \otimes X_k = \begin{pmatrix} 1 & 0 \\ 0 & -1 \end{pmatrix} \otimes \begin{pmatrix} 0 & 1 \\ 1 & 0 \end{pmatrix} \]
The matrix exponential is \[e^{-i x_j x_k Z_j X_k} = \cos(x_j x_k) I \otimes I - i \sin(x_j x_k) Z_j \otimes X_k.\] The full PauliFeatureMap is \[U_{Pauli}(\vec{x}) = \prod_{j=1}^n \left( e^{-i x_j X_j} e^{-i x_j Y_j} e^{-i x_j Z_j} \right) \prod_{j < k} e^{-i x_j x_k Z_j X_k}.\]
Fig.~\ref{pfmap} shows the PauliFeatureMap circuit, where each qubit undergoes an initial Hadamard gate followed by two rotations around the Z-axis (\(R_z(\theta_j)\)) and two rotations around the X-axis (\(R_x(\pm\pi/2)\)). The circuit ends with controlled-X (CNOT) gates that introduce entanglement between qubits, encoding complex feature interactions.

\subsection{Quantum Support Vector Classifier (QSVC)}
The QSVC~\cite{rebentrost2014quantum,gentinetta2024complexity} is an algorithm (Algorithm~\ref{qsvcAlgorithm}) that uses the principles of quantum computing concepts to classify inputs and should theoretically be more computationally efficient than classical methods. It is a quantum analogous to the classical Support Vector Machine (SVM) that use kernel functions to represent non-linear features and relies on information from Kernel Methods and Quantum Mechanics. QSVC effectively computes inner products in a vast Hilbert space using quantum parallelism and interference, and it can differentiate any data that would be difficult to classify initially using classical techniques \cite{gentinetta2024complexity}.
 
\begin{figure}[t]
\begin{center}
\includegraphics[scale=0.737]{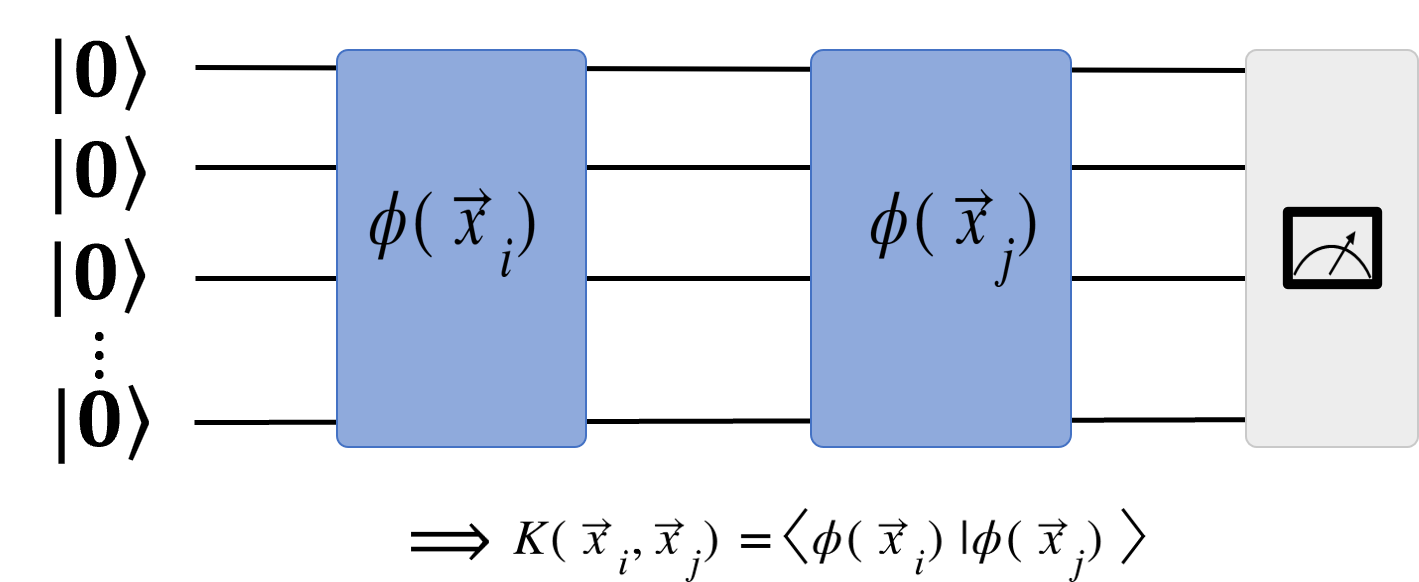}
\end{center}\vspace{-5 mm}
\caption{\rm Quantum circuit for QSVC.}
        \label{fig:qsvc_circuit} 
\end{figure}

The goal of QSVC is similar to classical SVM which is to find the hyperplane that optimally separates two data classes with the maximum margin. Given a classical dataset $\{(\vec{x}_i, y_i)\}$ where $\vec{x}_i \in \mathbb{R}^n$ are the genome sequence and $y_i \in \{0, 1\}$ are corresponding class labels, the separating hyperplane can be given as $\vec{w} \cdot \vec{x} + b = 0$.
To ensure that the hyperplane separates the data effectively, we use specific constraints given by eq. \ref{con_eq}:
 \begin{equation}\label{con_eq}
y_i (\vec{w} \cdot \vec{x}_i + b) \geq 1 \quad \forall i 
\end{equation}
The margin is the distance between the hyperplane and the closest data points, is $\frac{2}{\|\vec{w}\|}$. We aim to maximize the margin, equivalent to minimizing $\|\vec{w}\|$. Therefore the primal optimization problem is given by eq. ~\ref{opti_eq} and subject to eq. ~\ref{con_eq}.
\begin{equation}\label{opti_eq}
\min_{\vec{w}, b} \frac{1}{2} \|\vec{w}\|^2
\end{equation}
In QSVC, each data sample \(\vec{x}_i\) is encoded into a quantum state \(|\phi(\vec{x}_i)\rangle\) using a quantum feature map \( U_{\text{encode}}(\vec{x}) \) \cite{suzuki2024quantum}. 
The kernel matrix \(K\) \cite{zhou2024quantum} is then constructed by measuring the overlap, or inner product, of these quantum states and given by eq.  \ref{kernel_svm} : \begin{equation}\label{kernel_svm}
K(\vec{x}_i, \vec{x}_j) = \langle \phi(\vec{x}_i) | \phi(\vec{x}_j) \rangle
\end{equation} 
where \(K\) is essential for capturing the similarities between different data points in the transformed quantum space \cite{gentinetta2024complexity,jager2023universal}. The circuit depicted in Fig. \ref{fig:qsvc_circuit} initializes the quantum states, applies the kernel function, and measures the output state.
The optimization step involves solving the dual problem of the SVM. We introduce Lagrange multipliers $\alpha_i \geq 0$ for each constraint \cite{suzuki2024quantum}. Using eq. ~\ref{con_eq} and \ref{opti_eq} the Lagrangian is given by eq. ~\ref{Lagrang1}
\begin{equation}\label{Lagrang1}
\mathcal{L}(\vec{w}, b, \vec{\alpha}) = \frac{1}{2} \|\vec{w}\|^2 - \sum_{i=1}^n \alpha_i [y_i (\vec{w} \cdot \vec{x}_i + b) - 1]
\end{equation}
We need to find the saddle point of the Lagrangian, which involves minimizing $\mathcal{L}$ concerning $\vec{w}$ and $b$, and while maximizing it for $\vec{\alpha}$. Taking the partial derivatives of $\mathcal{L}$ to $\vec{w}$ and $b$ and setting them to zero.
Now substitute $\vec{w} = \sum_{i=1}^n \alpha_i y_i \vec{x}_i$ into the Lagrangian and
simplifying it, we get eq. ~\ref{Lagrangian}:
\begin{equation}\label{Lagrangian}
\mathcal{L}(\vec{\alpha}) = \sum_{i=1}^n \alpha_i - \frac{1}{2} \sum_{i,j=1}^n \alpha_i \alpha_j y_i y_j \vec{x}_i \cdot \vec{x}_j.
\end{equation}
Eq. ~\ref{Lagrangian} leads to the dual problem, where regularization parameter $C$ limits the upper bound of the Lagrange multipliers given by eq.  ~\ref{dual_opti_eq}:

\begin{algorithm}
\caption{Extended QSVC}
\label{qsvcAlgorithm}
\begin{algorithmic}[1]
\Procedure{Initialize}{}
\State Set $C$.
\EndProcedure
\Procedure{Quantum State Preparation}{$\vec{x}_i$}
\State Encode $\vec{x}_i$ via $U_{\text{encode}}(\vec{x}_i)$.
\EndProcedure
\Procedure{Quantum Kernel Matrix }{}
\State  $K_{ij} \leftarrow \langle \phi(\vec{x}_i) | \phi(\vec{x}_j) \rangle$.
\EndProcedure
\Procedure{Optimize SVM}{}
\State Solve:
\[
\max_{\vec{\alpha}} \left( \sum_{i=1}^n \alpha_i - \frac{1}{2} \sum_{i,j=1}^n y_i y_j \alpha_i \alpha_j K_{ij} \right)
\]
\State Subject to: $0 \leq \alpha_i \leq C$, $\sum \alpha_i y_i = 0$.
\EndProcedure
\Procedure{Compute Bias}{$b$}
\State $b = y_s - \sum_{i} \alpha_i y_i K_{is}$, for $0 < \alpha_s < C$.
\EndProcedure
\Procedure{Prediction}{$\vec{x}$}
\State Compute $f(\vec{x}) = \sum \alpha_i y_i K(\vec{x}_i, \vec{x}) + b$.
\State \textbf{return} $\text{sign}(f(\vec{x}))$.
\EndProcedure
\end{algorithmic}
\end{algorithm}
\begin{equation}\label{dual_opti_eq}
\max_{\vec{\alpha}} \sum_{i=1}^n \alpha_i - \frac{1}{2} \sum_{i,j=1}^n \alpha_i \alpha_j y_i y_j K(\vec{x}_i, \vec{x}_j)
\end{equation}
subject to: $\sum_{i=1}^n \alpha_i y_i = 0, \quad  0 \leq \alpha_i \leq C $.
This dual problem is a convex quadratic programming problem. Standard solvers like Sequential Minimal Optimization (SMO) \cite{platt1998sequential, jui2021performance} can be used to find the optimal $\alpha_i$. Once the optimal $\alpha_i$ is found, we compute:
\begin{equation}
\vec{w} = \sum_{i=1}^n \alpha_i y_i \phi(\vec{x}_i)
\end{equation}
Then, the decision function for a new data point $\vec{x}$ is given by 
\begin{equation}\label{decisionFuncQSVC}
f(\vec{x}) = \vec{w} \cdot \phi(\vec{x}) + b
\end{equation}
Since $\vec{w}$ is known in eq. ~\ref{decisionFuncQSVC}, we must find the bias term $b$. For any support vector $\vec{x}_i$, the constraint $y_i (\vec{w} \cdot \phi(\vec{x}_i) + b) = 1$ holds. Hence, $b$ can be computed as:
\begin{equation}\label{b_eq}
b = y_i - \sum_{j=1}^n \alpha_j y_j K(\vec{x}_j, \vec{x}_i)
\end{equation}
From eq. ~\ref{decisionFuncQSVC} and eq. ~\ref{b_eq} the final $f(\vec{x})$ becomes eq. ~\ref{final_dec_eq_qsvc}
\begin{equation}\label{final_dec_eq_qsvc}
f(\vec{x}) = \sum_{i=1}^n \alpha_i y_i K(\vec{x}_i, \vec{x}) + b
\end{equation}
 and the sign of $f(\vec{x})$ determines the class label, $\hat{y} = \text{sign}(f(\vec{x})).$

\subsection{Pegasos-QSVM}
The Pegasos-QSVC \cite{shalev2007pegasos} is a hybrid algorithm (Algorithm~\ref{pegasosQSVCAlgorithm}) that combines the principles of quantum computing with the Pegasos (Primal Estimated sub-GrAdient SOlver for SVM) optimization technique. It uses the Pegasos algorithm based on stochastic gradient descent for optimization and is intended to exploit quantum computing's favourable scaling with dimensionality. The goal of Pegasos-QSVC is to preserve computational effectiveness as well as accuracy for classification problems by combining quantum feature maps, quantum kernel techniques, and traditional optimization approaches
\cite{havlivcek2019supervised}.

\begin{algorithm}
\caption{Improved Pegasos-QSVC}
\label{pegasosQSVCAlgorithm}
\begin{algorithmic}[1]
\Procedure{Initialize}{}
\State Init $\vec{\theta}$, $\eta$, $\lambda$, $T$.
\EndProcedure
\Procedure{Quantum State Preparation}{$\vec{x}_i$}
\State Encode $\vec{x}_i$ via $U_{\text{encode}}(\vec{x}_i)$.
\EndProcedure
\Procedure{Quantum Kernel Matrix}{}
\State $K_{ij} \gets \langle \phi(\vec{x}_i) | \phi(\vec{x}_j) \rangle$.
\EndProcedure
\Procedure{Train}{}
\For{$t = 1$ to $T$}
\State Select subset $\{\vec{x}_i\}$.
\State For each $\vec{x}_i$, compute $y_i \langle \vec{w}, \phi(\vec{x}_i) \rangle$.
\If{$y_i \langle \vec{w}, \phi(\vec{x}_i) \rangle < 1$}
\State $\vec{w} \gets (1 - \eta \lambda) \vec{w} + \eta y_i \phi(\vec{x}_i)$.
\Else
\State $\vec{w} \gets (1 - \eta \lambda) \vec{w}$.
\EndIf
\State $\vec{w} \gets \min(1, \frac{1/\sqrt{\lambda}}{\|\vec{w}\|}) \vec{w}$.
\EndFor
\EndProcedure
\Procedure{Prediction}{$\vec{x}$}
\State Prepare $U_{\text{encode}}(\vec{x})$.
\State $f(\vec{x}) \gets \text{sign}(\langle \vec{w}, \phi(\vec{x}) \rangle)$.
\State \Return $\text{sign}(f(\vec{x}))$.
\EndProcedure
\end{algorithmic}
\end{algorithm}

The Pegasos-QSVC algorithm (Algorithm~\ref{pegasosQSVCAlgorithm}) operates through several key steps. Initially, necessary parameters are set up for quantum gates, including the initialization of the quantum gate parameters \(\vec{\theta}\), learning rate \(\eta\), regularization parameter \(\lambda\), and the number of iterations \(T\). Each data sample \(\vec{x}_i\) is then encoded into a quantum state using a quantum feature map \(U_{\text{encode}}(\vec{x}_i)\) \cite{schuld2015introduction}. This quantum state preparation transforms the classical data into quantum states, facilitating the computation of the quantum kernel matrix \(K\) through the overlap of these quantum states. The kernel matrix \(K_{ij}\) is computed as eq. ~\ref{kernel_svm} and 
the primal problem in Pegasos-QSVC with a quantum kernel is given by eq. ~\ref{eq:primal_problem} \cite{rebentrost2014quantum},
\begin{equation}
\min_{\vec{w}} \frac{\lambda}{2} \|\vec{w}\|^2 + \frac{1}{m} \sum_{i=1}^m \max(0, 1 - y_i (\vec{w} \cdot \phi(\vec{x}_i))).
\label{eq:primal_problem}
\end{equation}
Given the quantum kernel, the weight vector \(\vec{w}\) is expressed in the quantum feature space, making direct manipulation challenging. Instead, we work with the dual formulation or implicitly use the kernel in the Pegasos updates. In each iteration of Pegasos, we compute the sub-gradient based on a single sample \( (x_t, y_t) \) \cite{shalev2007pegasos}:
\begin{equation}
\nabla_{\vec{w}} f(\vec{w}) = \lambda \vec{w} + \frac{1}{m} \sum_{i: y_i (\vec{w} \cdot \phi(\vec{x}_i)) < 1} -y_i \phi(\vec{x}_i)
\label{eq:subgradient}
\end{equation}
Using the quantum kernel, the sub-gradient calculation involves evaluating the quantum kernel \( K(x_t, x_i) \) for all support vectors \( x_i \) and updating the weight vector using the computed kernel values \cite{biamonte2017quantum}.

During the training phase, for tractability~\cite{shalev2007pegasos}, the algorithm iteratively updates the weight vector \(\vec{w}\) using stochastic gradient descent. For each iteration, a subset of data samples is randomly selected, and the decision value \(y_i \langle \vec{w}, \phi(\vec{x}_i) \rangle\) is computed. If the decision value is less than 1, the weight vector is updated as follows:
\begin{equation}
\vec{w} \leftarrow (1 - \eta \lambda) \vec{w} + \eta y_i \phi(\vec{x}_i)
\label{eq:weight_update_1}
\end{equation}
Otherwise, the weight vector is updated as:
\begin{equation}
\vec{w} \leftarrow (1 - \eta \lambda) \vec{w}
\label{eq:weight_update_2}
\end{equation}
To ensure regularization, the weight vector \(\vec{w}\) is normalized:
\begin{equation}
\vec{w} \leftarrow \min\left(1, \frac{1/\sqrt{\lambda}}{\|\vec{w}\|}\right) \vec{w}
\label{eq:normalization}
\end{equation}
For making predictions, the algorithm computes the quantum state for a new data sample \(\vec{x}\) using \(U_{\text{encode}}(\vec{x})\). The decision function is then calculated as:
\begin{equation}
f(\vec{x}) = \text{sign}(\langle \vec{w}, \phi(\vec{x}) \rangle)
\label{eq:decision_function}
\end{equation}
Based on the sign of the decision function, the class label of the new data sample is determined.

\subsection{Variational Quantum Classifier (VQC)}

The Variational Quantum Classifier (VQC) \cite{jager2023universal} utilizes parameterized quantum circuits, which are optimized using classical algorithms to perform classification tasks efficiently\cite{havlivcek2019supervised}. This approach combines quantum computing capabilities with classical optimization to handle complex datasets effectively \cite{schuld2018supervised}.  

\begin{figure}[t]
\begin{center}
\includegraphics[scale=0.737]{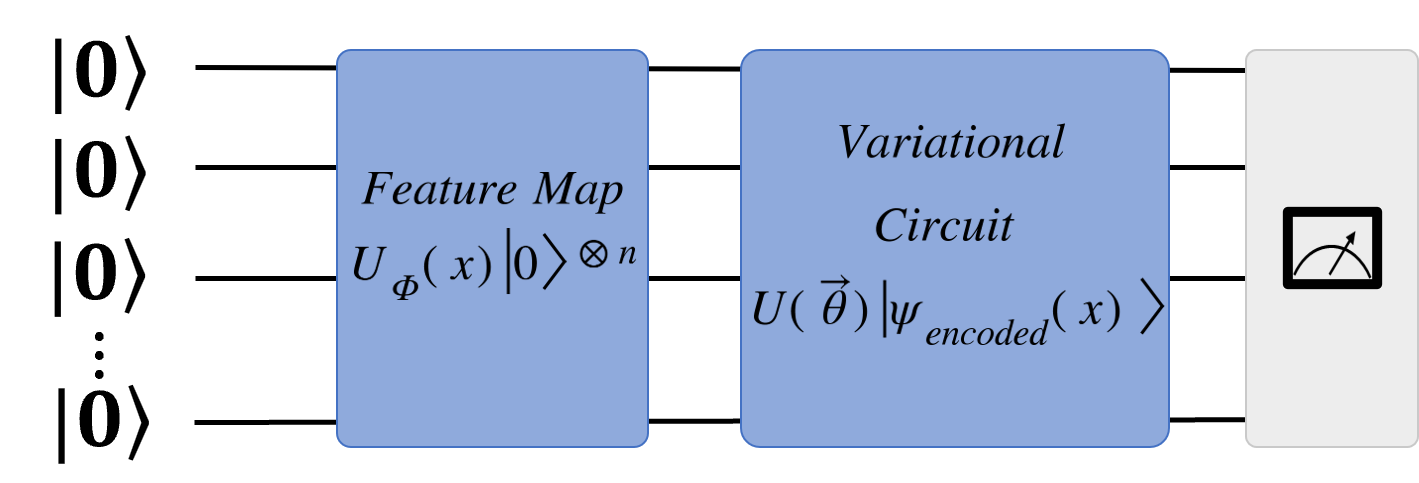}
\end{center}\vspace{-4mm}
\caption{\rm Quantum circuit for VQC.}
        \label{obj}
\end{figure}

\begin{algorithm}
\caption{Enhanced VQC}
\label{vqcAlgorithm}
\begin{algorithmic}[1]
\Procedure{Init}{}
\State Prepare $|0\rangle^{\otimes n}$ $\&$ Init $\vec{\theta}$.
\EndProcedure
\Procedure{Encode}{$x$}
\State $|\phi_{\text{encoded}}(x)\rangle \gets U(x) |0\rangle^{\otimes n}$.
\EndProcedure
\Procedure{Variational Circuit}{$x$, $\vec{\theta}$}
\State $|\phi(\vec{\theta}, x)\rangle \gets U(\vec{\theta}) |\phi_{\text{encoded}}(x)\rangle$.
\EndProcedure
\Procedure{Measure}{}
\State Measure $|\phi(\vec{\theta}, x)\rangle$.
\State $C(\vec{\theta}) \gets \frac{1}{N} \sum (y_i - \langle O \rangle_{\vec{\theta}, x_i})^2$.
\EndProcedure
\Procedure{Optimize}{}
\State $\frac{\partial \langle O \rangle}{\partial \vec{\theta}_i} \gets \frac{\langle O \rangle_{\vec{\theta}_i + \frac{\pi}{2}} - \langle O \rangle_{\vec{\theta}_i - \frac{\pi}{2}}}{2}$.
\While{not converged}
\State Update $\vec{\theta}$.
\EndWhile
\State \Return $\vec{\theta}_{\text{opt}}$.
\EndProcedure
\Procedure{Predict}{$x$, $\vec{\theta}_{\text{opt}}$}
\State $|\phi_{\text{opt}}(x)\rangle \gets U(\vec{\theta}_{\text{opt}}) U(x) |0\rangle^{\otimes n}$.
\State $f(x) \gets \text{sign}(\langle O \rangle_{\vec{\theta}_{\text{opt}}, x})$.
\State \Return $f(x)$.
\EndProcedure
\end{algorithmic}
\end{algorithm}

The VQC algorithm (Algorithm~\ref{vqcAlgorithm}) initiates by preparing the qubits in the initial state 
\( |\phi\rangle = |0\rangle^{\otimes n} \)), where \(n\) is the number of qubits. The variational parameters \(\vec{\theta}\) for the quantum gates are initialized. Classical data \(x\) is then encoded into this quantum state using an encoding unitary \(U_{\Phi}(x)\), transforming the initial state to \( |\phi_{\text{encoded}}(x)\rangle = U(x) |\phi_0\rangle \), laying the foundation for the variational circuit\cite{schuld2018supervised}.
The variational circuit is then constructed with parameterized quantum gates \(U(\vec{\theta})\). These gates are applied to the encoded quantum state \( |\phi_{\text{encoded}}(x)\rangle \), transforming it according to the variational parameters\cite{mcclean2018barren}:
\begin{equation}
|\phi(\vec{\theta}, x)\rangle = U(\vec{\theta}) |\phi_{\text{encoded}}(x)\rangle
\end{equation}
The transformed quantum state \( |\phi(\vec{\theta}, x)\rangle \) is measured in the computational basis. These outcomes are crucial as they are used to compute the cost function \(C(\vec{\theta})\), which is defined as eq. ~\ref{costfunc:vqc}, which quantifies the performance of the classifier and \(O\) represents the observable whose expectation value we compute  \cite{mitarai2018quantum,havlivcek2019supervised}.
\begin{equation}\label{costfunc:vqc}
C(\vec{\theta}) = \langle \phi(\vec{\theta}, x) | O | \phi(\vec{\theta}, x) \rangle 
\end{equation}
\begin{equation}
C(\vec{\theta}) = \frac{1}{N} \sum_{i=1}^{N} \left(y_i - \langle O \rangle_{\vec{\theta}, x_i}\right)^2
\end{equation}

The variational parameters \(\vec{\theta}\) are optimized using a classical optimization algorithm. The goal is to minimize the cost function \(C(\vec{\theta})\), which ideally corresponds to maximizing the classifier's accuracy. Optimization often employs gradient-based methods where gradients can be estimated using the parameter-shift rule\cite{mitarai2018quantum}:
\begin{equation}
\frac{\partial \langle O \rangle}{\partial \vec{\theta}_i} = \frac{\langle O \rangle_{\vec{\theta}_i + \frac{\pi}{2}} - \langle O \rangle_{\vec{\theta}_i - \frac{\pi}{2}}}{2}
\end{equation}

This optimization process continues until convergence, at which point the optimized parameters \(\vec{\theta}_{\text{opt}}\) are obtained\cite{temme2017error}:
\begin{equation}
\vec{\theta}_{\text{opt}} = \arg \min_{\vec{\theta}} C(\vec{\theta})
\end{equation}
Once the parameters \(\vec{\theta}\) are optimized, the VQC can classify new inputs. For a given new input \(x\), the quantum circuit is configured with the optimized parameters \(\vec{\theta}_{\text{opt}}\), and the state \( |\phi_{\text{opt}}(x)\rangle \) is prepared and measured:
\[ f(x) = \text{sign}(\langle \phi_{\text{opt}}(x) | O | \phi_{\text{opt}}(x) \rangle) \]
The \(f(x)\) determines the class label of the input \(x\) based on the sign of the expectation value of the observable \(O\).

\subsection{Quantum Neural Network (QNN)}

A QNN \cite{abbas2021power} is a computational model that integrates the principles of quantum computing with the architecture of classical neural networks. It is a specific adaption of VQC. QNN is a quantum analog of classical neural networks that uses the inherent parallelism and entanglement aspects of quantum mechanics for performing complex computations significantly quicker than traditional techniques. QNNs process information in ways orthogonal to classical computers through the use of qubits and quantum gates, with broad applications across disciplines like machine learning optimization, as well as within computational pipelines dedicated towards general-purpose quantum computation \cite{abbas2021power}.

The operation of a QNN (Algorithm~\ref{qnnAlgorithm}) involves several stages. Initially, qubits are prepared in the state \( |\phi_0\rangle = |0\rangle^{\otimes n} \), where \( n \) is the number of qubits. The parameters \( \theta \) for the parameterized quantum gates are then initialized. Each classical data point \( x \) is encoded into a quantum state using an encoding unitary operation \( U_{\text{encode}}(x) \) \cite{schuld2018supervised}. 

The core of the QNN consists of multiple quantum layers, each applying a unitary transformation \( U_l(\theta_l) \) parameterized by \( \theta_l \). These transformations represent the quantum analog of classical neural network layers and are designed to learn complex patterns in the data\cite{biamonte2017quantum}. 
Entanglement between qubits is introduced using CNOT gates, creating correlations between different parts of the quantum system that are crucial for capturing complex dependencies in the data\cite{mcclean2016theory}.

\begin{algorithm}
\caption{Advanced QNN}
\label{qnnAlgorithm}
\begin{algorithmic}[1]
\Procedure{Init}{}
\State Prepare $|0\rangle^{\otimes n}$ $\&$ Init $\theta$.
\EndProcedure

\Procedure{Encode}{$x$}
\State $|\phi_{\text{enc}}(x)\rangle \gets U_{\text{encode}}(x) |\phi_0\rangle$.
\EndProcedure

\Procedure{Apply Layers}{}
\For{$l = 1$ to $L$}
\State $|\phi_l\rangle \gets U_l(\theta_l) |\phi_{l-1}\rangle$.
\State Entangle with CNOT gates.
\EndFor
\EndProcedure

\Procedure{Measure}{}
\State Measure $|\phi_L\rangle \to |\Phi\rangle$.
\State $E \gets |\text{expected} - \langle \Phi | O | \Phi \rangle|^2$.
\EndProcedure

\Procedure{Optimize}{}
\While{not converged}
\State $\nabla_{\theta} E \gets \frac{E(\theta + \frac{\pi}{2}) - E(\theta - \frac{\pi}{2})}{2}$.
\State $\theta \gets \theta - \alpha \nabla_{\theta} E$.
\EndWhile
\EndProcedure
\Procedure{Predict}{$x$, $\theta_{\text{opt}}$}
\State $|\Phi_{\text{opt}}\rangle \gets U_L(\theta_{\text{opt},L}) \ldots U_1(\theta_{\text{opt},1}) |\phi_{\text{enc}}(x)\rangle$.
\State \Return $\text{sign}(\langle \Phi_{\text{opt}} | O | \Phi_{\text{opt}} \rangle)$.
\EndProcedure
\end{algorithmic}
\end{algorithm}

After the quantum transformations, the qubits are measured to obtain the final quantum state \( |\Phi\rangle \). The measurement outcomes are used to compute the error \( E \) between the expected output and the actual measurement \cite{abbas2021power}. This error is minimized using a quantum version of gradient descent, where the parameters \( \theta \) are updated iteratively:
\begin{equation}
\theta = \theta - \alpha \nabla_{\theta} E
\end{equation}
where, \( \alpha \) denotes the learning rate, while \( \nabla_{\theta} E \) signifies the gradient of the error concerning the parameters ~\cite{ruder2016overview}.

The mathematical formulation of a QNN involves several key components. The initial state preparation and data encoding can be represented   as:
\begin{equation}
|\phi_{\text{encoded}}(x)\rangle = U_{\text{encode}}(x) |\phi_0\rangle
\end{equation}

Each layer of the quantum neural network applies a unitary transformation:
\begin{equation}
|\phi_l(\theta_l)\rangle = U_l(\theta_l) |\phi_{l-1}(\theta_{l-1})\rangle
\end{equation}
where \( l \) denotes the layer index, and \( \theta_l \) are the parameters for the \( l \)-th layer. 
The final state after all layers and entanglement operations is measured to obtain the output state \( |\Phi\rangle \):
\begin{equation}
|\Phi\rangle = U_L(\theta_L) \ldots U_1(\theta_1) |\phi_{\text{encoded}}(x)\rangle
\end{equation}
The error \( E \) is computed based on the difference between the expected output and the measurement result:
\begin{equation}
E = \left| \text{expected\_output} - \langle \Phi | O | \Phi \rangle \right|^2
\end{equation}
If the observable \( O \) corresponds to a measurement in the computational basis (e.g., \( O = Z \)), the expectation value can be written as:
\begin{equation}
\langle \Phi | O | \Phi \rangle = \langle \Phi | Z | \Phi \rangle
\end{equation}

The optimization of the parameters \( \theta \) is performed using quantum gradient descent, which iteratively updates the parameters to minimize the error. The gradient can be estimated using the parameter-shift rule\cite{schuld2019evaluating}:
\begin{equation}
\frac{\partial E}{\partial \theta_i} = \frac{E(\theta_i + \frac{\pi}{2}) - E(\theta_i - \frac{\pi}{2})}{2}
\end{equation}
Updating the parameters using gradient descent:
\begin{equation}
\theta_i = \theta_i - \alpha \frac{\partial E}{\partial \theta_i}
\end{equation}
where \( \alpha \) is the learning rate\cite{ruder2016overview}.
Once the parameters are optimized, the QNN can be used to classify new inputs. The decision function uses the optimized parameters \( \theta_{\text{opt}} \) to determine the class label. For a new input \( x \), the encoded state is:
\begin{equation}
|\phi_{\text{encoded}}(x)\rangle = U_{\text{encode}}(x) |\phi_0\rangle
\end{equation}
Applying the optimized variational circuit:
\begin{equation}
|\Phi_{\text{opt}}\rangle = U_L(\theta_{\text{opt},L}) \ldots U_1(\theta_{\text{opt},1}) |\phi_{\text{encoded}}(x)\rangle
\end{equation}
The decision function is then:
\begin{equation}
f(x) = \text{sign}(\langle \Phi_{\text{opt}} | O | \Phi_{\text{opt}} \rangle)
\end{equation}
where \( \text{sign}(y) \) is a function that returns \( +1 \) if \( y \geq 0 \) and \( -1 \) if \( y < 0 \). This determines the class label based on whether the measured expectation value is positive or negative.

\section{Convergence Analysis}


In this section, we examine the convergence of the previously  developed and extended QML algorithms. 

The analysis of the proposed QSVM below includes the convexity of the optimization problem, the existence and uniqueness of the solution, the the Karush-Kuhn-Tucker (KKT) conditions to understand the convergence and the accuracy of quantum kernel estimation. The dual form of the QSVM optimization problem and kernel function is given by eq. ~\ref{dual_opti_eq} and eq. ~\ref{kernel_svm}
From eq. ~\ref{dual_opti_eq}, the objective function is a quadratic function of $\vec{\alpha}$. The matrix $P$ in the quadratic term $\frac{1}{2} \vec{\alpha}^T P \vec{\alpha}$ is given by $P_{ij} = y_i y_j K(\vec{x}_i, \vec{x}_j)$. Since the kernel matrix $K$ is positive semi-definite, the matrix $P$ is also positive semi-definite. Therefore, the objective function is concave, and the problem is a convex optimization problem \cite{suzuki2024quantum}.

Given that the dual problem is a convex quadratic programming problem with linear constraints, a global maximum exists. The solution is unique if the kernel matrix $K$ is positive definite. If $K$ is only positive semi-definite, multiple solutions may exist, but any solution will be optimal. The Karush-Kuhn-Tucker (KKT) conditions establish the necessary and sufficient criteria for achieving an optimal solution in a convex optimization problem\cite{rebentrost2014quantum}.
Primal feasibility is expressed as:
\begin{equation}
0 \leq \alpha_i \leq C \quad \forall i
\end{equation}
Dual feasibility is given by:
\begin{equation}
\sum_{i=1}^n \alpha_i y_i = 0
\end{equation}

The Lagrangian for the dual problem is calculated as eq. ~\ref{Lagrangian}. Now
taking the partial derivative with respect to $\alpha_i$ and setting it to zero, we get:
\begin{equation}
\frac{\partial \mathcal{L}}{\partial \alpha_i} = 1 - \sum_{j=1}^n \alpha_j y_j y_i K(\vec{x}_i, \vec{x}_j) = 0 \quad \forall i
\end{equation}
This gives the stationarity condition:
\begin{equation}
1 - \sum_{j=1}^n \alpha_j y_j K(\vec{x}_i, \vec{x}_j) = 0
\end{equation}
Complementary slackness ensures that if a constraint is not active, the corresponding Lagrange multiplier must be zero. For the QSVM, this condition is:
\begin{equation}
\alpha_i \left( y_i \left( \sum_{j=1}^n \alpha_j y_j K(\vec{x}_j, \vec{x}_i) + b \right) - 1 \right) = 0
\end{equation}
As we mentioned in subsection 2.2, SMO breaks the problem into smaller sub-problems involving pairs of Lagrange multipliers. First, initialize with $\alpha_i = 0$ for all $i$. Iterate until convergence by selecting pairs $(\alpha_i, \alpha_j)$ to optimize. Solve the optimization sub-problem for these pairs while keeping other $\alpha_k$ (for $k \neq i, j$) fixed. The sub-problem is given by:
\begin{equation}
\max_{\alpha_i, \alpha_j} \left( \alpha_i + \alpha_j - \frac{1}{2} (y_i \alpha_i + y_j \alpha_j)^2 K(\vec{x}_i, \vec{x}_j) \right)
\end{equation}
Update $\alpha_i$ and $\alpha_j$ using the constraints: $0 \leq \alpha_i \leq C, \quad 0 \leq \alpha_j \leq C$ 
Finally, after each update, check if the KKT conditions are satisfied. The algorithm converges when all KKT conditions are met within a specified tolerance\cite{jui2021performance}.

The convergence analysis of the Pegasos algorithm below illustrates its efficiency and effectiveness. The algorithm, which performs stochastic gradient descent on the primal objective with a carefully chosen step size, demonstrates that the number of iterations needed to achieve an accuracy of \(\epsilon\) is \(O\left(\frac{1}{\lambda \epsilon}\right)\). This rate is notably superior to other stochastic gradient descent methods that typically require \(\Omega\left(\frac{1}{\epsilon^2}\right)\) iterations \cite{shalev2007pegasos}.

To analyze convergence, we consider the average objective of the algorithm compared to that of the optimal solution, denoted by \(w^* = \arg \min_w f(w)\). The main theorem states:
\begin{equation}
\frac{1}{T} \sum_{t=1}^T f(w_t) \leq f(w^*) + \frac{G^2 (1 + \ln(T))}{2 \lambda T}
\label{eq:main_theorem}
\end{equation}
where \(G\) is a bound on the sub-gradient norms, and \(T\) is the number of iterations with \(T \geq 3\).

Due to the convexity of \(f\):
\begin{equation}
f\left(\frac{1}{T} \sum_{t=1}^T w_t\right) \leq \frac{1}{T} \sum_{t=1}^T f(w_t)
\label{eq:convexity}
\end{equation}
This inequality, along with the main theorem, helps derive the deterministic convergence analysis when all data points are used in each iteration.

The step size \(\eta_t = \frac{1}{\lambda t}\) ensures convergence to the optimal solution at a rate of \(O\left(\frac{1}{\lambda \epsilon}\right)\). To see why this is the case, consider the following bound on the objective function value after \(T\) iterations:
\begin{equation}
f(w_T) - f(w^*) \leq \frac{\|w^*\|^2}{2 \eta_T T} + \frac{G^2}{2 \lambda T} \sum_{t=1}^T \eta_t
\label{eq:objective_bound}
\end{equation}
Given \(\eta_t = \frac{1}{\lambda t}\), we have:
\begin{equation}
\sum_{t=1}^T \eta_t = \frac{1}{\lambda} \sum_{t=1}^T \frac{1}{t} \approx \frac{1}{\lambda} \ln(T)
\label{eq:eta_sum}
\end{equation}
Substituting this back, we get:
\begin{equation}
f(w_T) - f(w^*) \leq \frac{\|w^*\|^2 \lambda}{2 T} + \frac{G^2 (1 + \ln(T))}{2 \lambda T}
\label{eq:final_bound}
\end{equation}
This confirms that \(T\) iterations are sufficient to achieve the desired accuracy \(\epsilon\).


For a thorough convergence analysis of the VQC below, we explore several crucial aspects related to the properties of the parameterized quantum circuits used, the computation of gradients, and overall convergence conditions that are unique to quantum computations.

The effectiveness of the VQC largely depends on the parameterized quantum circuit \( U(\vec{\theta}) \), which manipulates the quantum state that encodes classical data. The convergence of the VQC is deeply intertwined with the expressiveness and trainability of this circuit \cite{mcclean2018barren}. Expressiveness is determined by the circuit's ability to represent a broad class of quantum states needed for classification tasks. This expressiveness is quantified by how effectively \( U(\vec{\theta}) \) can explore the Hilbert space:
\begin{equation}
|\phi(\vec{\theta}, x)\rangle = U(\vec{\theta}) |\phi_{\text{encoded}}(x)\rangle
\end{equation}

However, a high level of expressiveness can lead to barren plateaus in the optimization landscape, where the gradient of the cost function to the parameters \(\vec{\theta}\) effectively vanishes:
\begin{equation}
\mathbb{E}\left[ \|\nabla_{\vec{\theta}} C(\vec{\theta})\|^2 \right] \sim e^{-\gamma n}
\end{equation}
Here, \( n \) denotes the number of qubits, and \( \gamma \) is a constant dependent on the circuit's architecture and depth. Additionally, the trainability of the circuit, which refers to the efficient updatibility of \(\vec{\theta}\) based on computed gradients, is crucial. The gradient, estimated using the parameter-shift rule, should be significant for effective training\cite{mitarai2018quantum}:
\begin{equation}
\frac{\partial C}{\partial \theta_i} = \frac{C(\vec{\theta} + \frac{\pi}{2} \hat{e}_i) - C(\vec{\theta} - \frac{\pi}{2} \hat{e}_i)}{2}
\end{equation}

Optimization of the VQC is to tune \(\vec{\theta}\) such that cost function  \( C(\vec{\theta}) \)tends to a minimum, usually by using variations of gradient descent more extensively. The parameter update rule in a typical gradient descent iteration is:
\begin{equation}
\vec{\theta}_{t+1} = \vec{\theta}_t - \eta \nabla_{\vec{\theta}} C(\vec{\theta}_t)
\end{equation}
where, \(\eta\) is the learning rate. The selection of \(\eta\) and the structure of this landscape in terms of local minima and saddle points is fundamental to selecting an appropriate numerical method as it can also influence how easily one could reach global minimum using these methods \cite{temme2017error}.
Stability of the VQC is determined through an analysis of the Hessian \( H_{\vec{\theta}} \) of the cost function defined as:
\begin{equation}
H_{ij}(\vec{\theta}) = \frac{\partial^2 C}{\partial \theta_i \partial \theta_j}
\end{equation}
A positive semi-definite Hessian implies a smooth optimization landscape with no sharp minima, thus leading to more reliable convergence\cite{schuld2018supervised}.

Convergence conditions for QNNs are a function of minimization of error/cost function \(E\) in an optimization process. This function calculates the difference between predicted outcomes and true outcomes, the reduction of which to a minimum is quite important for model performance.

The mathematical exploration begins with the behaviour of the cost function:
\begin{equation}
E(\theta) = \frac{1}{N} \sum_{i=1}^{N} \left( y_i - \langle \phi(\theta, x_i) | O | \phi(\theta, x_i) \rangle \right)^2
\end{equation}
where \(\theta\) denotes the variational parameters, \(y_i\) the actual outputs, \(x_i\) the input data, and \(O\) the observable.

Optimization typically employs gradient descent, updating parameters as follows:
\begin{equation}
\theta_{t+1} = \theta_t - \alpha \nabla_{\theta} E(\theta_t)
\end{equation}
Here, \(\alpha\) is the learning rate, and \(\nabla_{\theta} E(\theta_t)\) represents the gradient at iteration \(t\). Convergence is approached when the parameter updates become negligible:
\begin{equation}
\|\theta_{t+1} - \theta_t\| < \epsilon
\end{equation}
indicating a minimal change in parameters and potential convergence.
For gradient estimation, QNNs often utilize the parameter-shift rule, effectively computing gradients within quantum circuits\cite{schuld2019evaluating}:
\begin{equation}
\frac{\partial E}{\partial \theta_i} = \frac{E(\theta_i + \frac{\pi}{2}) - E(\theta_i - \frac{\pi}{2})}{2}
\end{equation}

The convergence rate depends on factors like learning rate, the curvature of the error function analyzed through its Hessian matrix, and the inherent noise in gradient estimates. The eigenvalues of the Hessian govern the optimal bounds for the learning rate's stability.

\section{Experiments and Evaluations}
\begin{figure}[h]
\begin{center}
\includegraphics[scale=0.340]{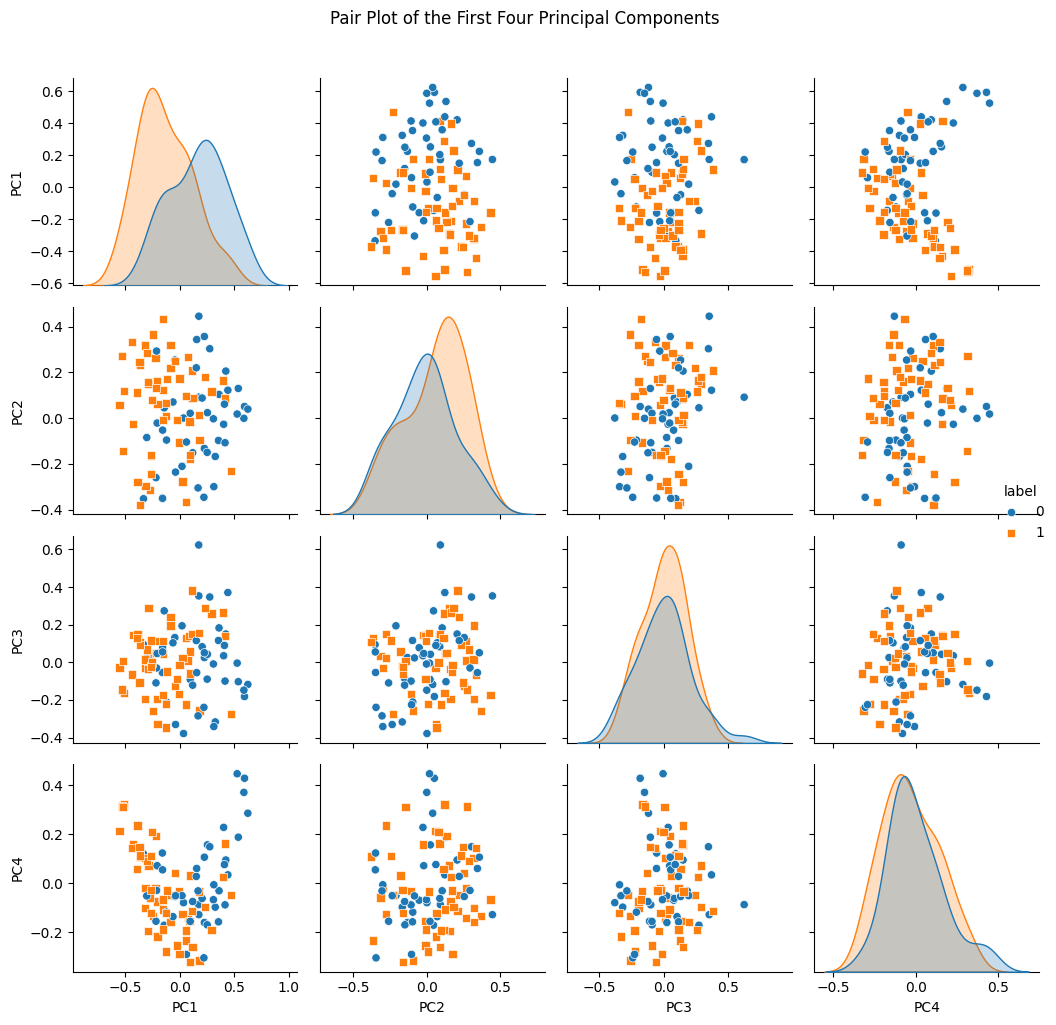}
\end{center}\vspace{-4 mm}
\caption{\rm Pair Plot of the First Four Principal Components for the First 100 Genome Sequence}
        \label{pca}
\end{figure}

To evaluate the performance of the aforementioned QML models, we utilize the Qiskit AerSimulator, which replicates the properties and behavior of real-world IBM quantum computers \cite{Qiskit}. In our experiments, four qubits are utilized to handle the benchmark genome sequence dataset \textit{democoding vs. intergenomic}~\cite{grevsova2023genomic} which contains 100,000 genome sequences with two classes of transcripts. To test the performance of the QML models, asubset of dataset is chosen and slip into training and testing sets. Initially, the genome sequences are converted into numerical format via text vectorization and then undergo dimensionality reduction using PCA, as illustrated in Fig. ~\ref{pca}. This figure presents a pair plot of the first four principal components (PCs) from a PCA on our dataset. The axes represent PC1, PC2, PC3, and PC4, capturing the maximum variance in the data. Scatter plots show relationships between pairs of PCs, while diagonal cells display density plots for each PC, illustrating class distribution. The colour-coding by labels highlights class separability in the principal component space. Feature mapping techniques prepare the data for encoding as quantum data, allowing for processing in QML models. This visualization indicates how PCA reduces dimensionality while maintaining critical information.


Fig.~\ref{obj} shows the convergence of the squared loss objective function for QNN and VQC using ZFeatureMap, ZZFeatureMap and PauliFeatureMap in the training phase. The squared loss function computes the difference between predicted and true values, which reduces in most instances exponentially (at least during the first training period). The most notable and continuous divergence from the objective function value is seen in the QNN plot for ZFeatureMap, suggesting improved relative performance. The ZZFeatureMap and PauliFeatureMap perform well and poorly, respectively. Similarly, in the VQC plot, ZFeatureMap has lower objective function values and faster convergence than other feature maps. This means that data representation is better for learning, which results in a quicker model. VQC models make use of the ZFeatureMap but are less prone to feature map selection.

\begin{figure}[h]
\begin{center}
\includegraphics[scale=0.330]{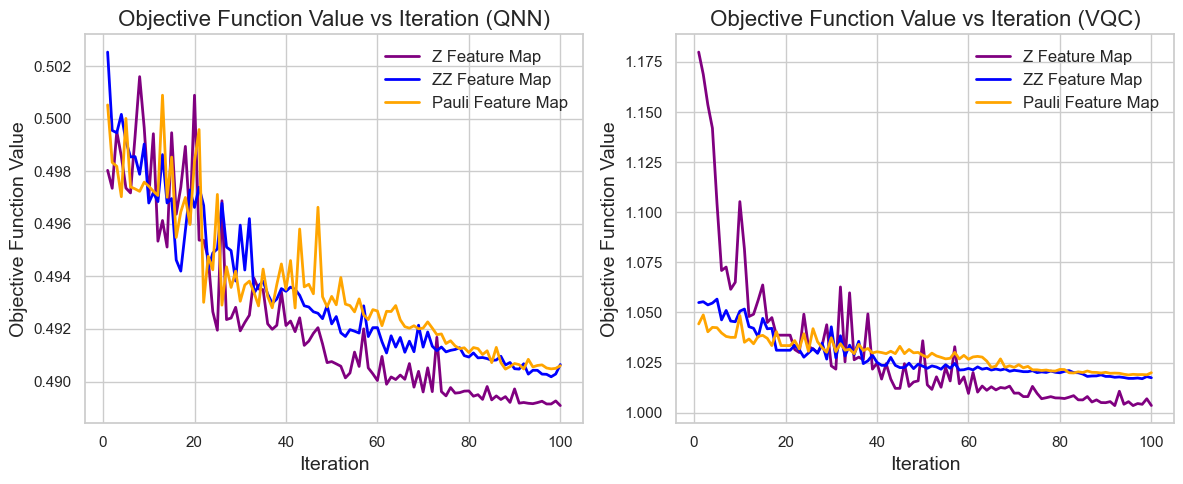}
\end{center}\vspace{-4mm}
\caption{Convergence of QNN \& VQC objective functions.}
\label{obj}
\end{figure}

 Table~\ref{res_tab} presents our comprehensive evaluation using classification metrics. It demonstrates a significant improvement in training and testing accuracies, and other performance matrices such as F1-Score, Area Under the Receiver Operating Characteristics (AUROC), recall, and precision, outperforming the results of ~\cite{abbas2021power}, ~\cite{gentinetta2024complexity}, \cite{jui2021performance}, and ~\cite{pokhrel2024quantum} in differentiating between two genome sequence classes.

\subsection{Observations and Discussion}
The results presented in Table~\ref{res_tab} demonstrate how performance is affected by the feature mapping and classification technique selection. Using the ZFeatureMap, the QSVM algorithm achieves 51.89\% training accuracy and 51.30\% test accuracy, with balanced but moderate precision, recall, F1 score, and AUROC (Table~\ref{res_tab}, row 1). With little overfitting or underfitting and modest classification performance, the ZFeatureMap most likely offers simple data representation. With a high recall of 99.12\% and an F1 score of 67.41\%, which indicates excellent sensitivity to positive classifications, the Peg-QSVM performs well despite having comparable accuracy rates (Table~\ref{res_tab}, row 2). The Peg-QSVM detects positive occurrences more accurately because of its algorithmic focus on maximizing recall.
The maximum recall-based algorithmic approach of Peg-QSVM makes it apt at more effective positive instance detection. The VQC shows 52.90\% training accuracy but lower test accuracy at 49.63\%, suggesting generalization issues  (Table~\ref{res_tab}, row 3). It is underfitted for the training set when it gives such results and is likely to underperform on test data encountered previously. QNN achieves the highest training accuracy (54.38\%) and decent test accuracy (51.65\%) (Table~\ref{res_tab}, row 4). It also has a high recall, indicating that the sensitivity is improved. The structure of QNN may let it to capture complicated data patterns more effectively, resulting in increased training accuracy and sensitivity to positive instances.

\begin{table*}[h]
\centering
\resizebox{\textwidth}{!}{
\begin{tabular}{l|l|l|l|l|l|l|l}
\multirow{2}{*}{\textbf{Feature Map}} & \multirow{2}{*}{\textbf{Our Algos}} & \multicolumn{1}{c|}{\textbf{Training}} & \multicolumn{5}{c}{\textbf{Test}} \\ \cline{3-8} 
 &  & \textbf{Accuracy} & \multicolumn{1}{l|}{\textbf{Accuracy}} & \multicolumn{1}{l|}{\textbf{Precision}} & \multicolumn{1}{l|}{\textbf{Recall}} & \multicolumn{1}{l|}{\textbf{F1}} & \textbf{AUROC} \\ \hline
\multirow{4}{*}{ZFeatureMap}  
 & QSVM & 51.89 & 51.30 & 51.61 & 52.71 & 52.16 & 50.01 \\ 
 & Peg-QSVM & 51.68 & 51.08 & 51.07 & 99.12 & 67.41 & 50.04 \\  
 & VQC & 52.90 & 49.63 & 50.61 & 54.46 & 52.47 & 49.52 \\  
 & QNN & 54.38 & 51.65 & 52.06 & 66.80 & 58.52 & 51.83 \\ \hline
\multirow{4}{*}{ZZFeatureMap} 
 & QSVM & 52.98 & 51.60 & 53.11 & 49.81 & 51.41 & 51.65 \\ 
 & Peg-QSVM & 52.44 & 51.02 & 51.04 & 99.76 & 67.53 & 49.98 \\  
 & VQC & 54.10 & 50.00 & 51.12 & 46.91 & 50.00 & 50.70 \\ 
 & QNN & 55.02 & 50.10 & 51.31 & 44.12 & 47.45 & 50.23 \\ \hline
\multirow{4}{*}{PauliFeatureMap} 
 & QSVM & 51.20 & 50.80 & 52.24 & 50.00 & 51.09 & 50.82 \\ 
 & Peg-QSVM & 52.06 & 51.05 & 51.30 & 99.41 & 67.46 & 50.01 \\ 
 & VQC & 52.10 & 50.55 & 51.60 & 50.49 & 51.04 & 50.55 \\ 
 & QNN & 53.20 & 50.52 & 51.67 & 47.75 & 49.63 & 50.58 \\ \hline
\end{tabular}
}
\caption{Performance Metrics for Various Quantum Feature Maps and Algorithms}\label{res_tab}
\end{table*}

Using ZZFeatureMap, QSVM achieves 52.98\% training accuracy and 51.60\% test accuracy (Table~\ref{res_tab}, row 5). ZZFeatureMap may provide a more accurate representation of data structure, hence improving QSVM training without compromising test performance. Peg-QSVM retains strong recall (99.76\%) and an F1 score (67.53\%), similar to ZFeatureMap performance, with fairly poor classification accuracy throughout testing phase (Table~\ref{res_tab}, row 6). The high recall indicates that the model is still sensitive to detecting positive cases, but poor test accuracy may imply overfitting. VQC provides low test accuracy and balanced metrics, therefore it may not be an appropriate solution for ZZFeatureMap (Table~\ref{res_tab}, row 7). As a result, the representation from ZZFeatureMap may not be properly exploited by VQC, resulting in underperformance.
QNN improves train accuracy to 55.02\% but decreases total test accuracy to 50.10\%, potentially indicating overfitting (Table~\ref{res_tab}, row 8). The QNN's complex architecture allows it to better identify patterns in training data, but it also causes overfitting, which is evident by lower test accuracy.

While using the PauliFeatureMap, QSVM has achieved the lowest training (51.2\%) and test accuracy (50.80\%) when comparing to other feature maps, showing balanced but average performance metrics (Table~\ref{res_tab}, row 9). The PauliFeatureMap might not represent the data structure as effectively or the representation from PauliFeatureMap is not properly exploited by QSVC, leading to lower accuracies. Peg-QSVM shows a high recall of 99.41\% and an F1 score of 67.46\%, indicating high sensitivity similar to other feature maps (Table~\ref{res_tab}, row 10). VQC shows balanced metrics with moderate test accuracy (Table~\ref{res_tab}, row 11). Although VQC appears to have balanced metrics, the test accuracy is moderate. The VQC shows consistent performance regardless of the feature map yet underperforms overall which suggests the general approach taken by this method in training is not fully exploiting activities that give optimal results with specific feature maps. QNN achieves slightly better training accuracy (53.2\%) with balanced precision and recall metrics, indicating moderate overall performance (Table~\ref{res_tab}, row 12). The QNN's architecture continues to model the patterns in training data well, but balanced precision and recall suggest it does not overfit towards extreme sensitivity or specificity.

The choice of feature map has a significant influence on the quantum classifier's performance. Though ZZFeatureMap is typically more accurate in training than ZFeatureMap and PauliFeatureMap, resulting in a better data structure representation for the quantum model, it does not necessarily increase test accuracy because of overfitting. 
The ZZFeatureMap allows training data to fit models, which could result in overfitting and impact the generalization of testing data. Among algorithms, Peg-QSVM typically achieves high recall and F1 scores, notably with ZFeatureMap and ZZFeatureMap, showing strong sensitivity to positive occurrences but potentially false positives, characterized by below-average accuracy. This pattern is most likely due to the Pegasos algorithm prioritizing recall above precision. The Pegasos algorithm in Peg-QSVM prioritizes recall, which aids in recognizing positives but may result in more false positives. VQC has consistent but low performance across feature maps, with reduced test accuracy indicating generalization issues, most likely related to optimizing variational parameters in high-dimensional quantum states. Because of the difficulty of optimizing parameters in quantum state space, the VQC technique may not be suitable to all applications. 
QNN has the best training accuracies across all feature maps, showing robust learning capabilities; nevertheless, test accuracy decreases suggesting potential overfitting. QNN's balanced precision and recall scores indicate more steady performance than other methods. The QNN's design effectively captures the complexities of training data but may overfit, despite balanced metrics indicating that it handles classification tasks more reliably than others.

\section{Concluding Remarks}
This paper comprehensively explores the application of QML algorithms for genomic data classification. It addresses key research questions, presents solutions, and provides detailed implementation and evaluation of various QML algorithms. The study demonstrates the potential of QML to enhance the efficiency and accuracy of genomic data analysis while identifying areas for further research and improvement.

Our initial analysis indicates that QML enhances genomic classification performance. Further studies with larger datasets will offer deeper insights \cite{10.1145/3663408.3665808}. The choice of feature map significantly affects quantum classifier performance. The ZZFeatureMap generally improves training accuracy compared to the ZFeatureMap and PauliFeatureMap, suggesting better data structure representation.

Peg-QSVM consistently shows high recall and F1 scores, making it suitable for detecting positives. QNN offers a balanced approach but may require overfitting mitigation. VQC exhibits consistent but moderate performance across feature maps, with generalization challenges due to the complexity of optimizing variational parameters.

Our findings highlight the promise of QML in genomic sequence classification. Selecting appropriate feature maps and algorithms is crucial to balancing sensitivity and specificity. Future work should address QNN overfitting and explore advanced feature mapping techniques to leverage the computational power of quantum devices, contributing to the broader application of QML in genomics and transforming genetic data analysis and personalized medicine. Additionally, QML models and feature mapping techniques are affected by inherent quantum noise and crosstalk. Therefore, we will evaluate the performance of these QML models and feature mapping techniques under such conditions to further assess their feasibility for genome sequence classification in both binary and multiclass tasks.

\bibliographystyle{ieeetr}
\bibliography{main}
\end{document}